\documentclass{article} 
\usepackage{iclr2016_conference,times}
\usepackage{hyperref}
\usepackage{url}
\usepackage{graphicx}
\usepackage{floatrow}
\usepackage{amssymb}

\title{Feature-based Attention in Convolutional Neural Networks}

\author{Grace W. Lindsay \\
Center for Theoretical Neuroscience\\
Department of Neuroscience\\
Columbia University\\
New York, NY 10032, USA \\
\texttt{gwl2108@columbia.edu} \\
}

\begin{document}

\maketitle

\begin{abstract}
-
\end{abstract}

\section{Introduction}

Attention is widely studied in neuroscience and becoming popular in deep learning as well, with particular applicability to image processing. While attention comes in many forms, spatial-based attention is most prominently featured in both fields. Feature-based attention, however, also engenders benefits for visual processing, and could be of use to artificial vision. Here, I briefly review some uses of attention in CNNs and introduce feature-based attention, which is a spatially-global alteration to a pre-trained CNN, applied in a category-specific way. More simply, using mechanisms inspired by biology, FBA applied to a given category works by biasing activity in the CNN towards the average activity pattern created by that category. This enhances detection of a given object class using a single feedforward pass of the CNN.

\section{Summary of Previous Work}
Previous work with CNNs has incorporated ideas from biological vision to enhance processing, especially with regards to biological attention. Taking inspiration from biology is reasonable, as CNNs have many architectural elements in common with the ventral visual stream, and visualization techniques have shown that feature representations are similar at corresponding levels of the CNN and visual stream ~\citep{zeiler2014visualizing}. Here, a recap of previous attention approaches in CNNs is provided, and background on biological feature attention is given. 

\subsection{Attention in CNNs}
A lot of work on attention in CNNs has focused on spatial attention. That is, performance is enhanced by processing small regions of the image in sequence (see ~\cite{mnih2014recurrent} for their work and their summary of previous work), or allocating processing resources according to spatial saliency maps ~\citep{xu2015show}. Biologically speaking, these "hard" and "soft" forms of attention in the machine learning literature correspond to "overt" and "covert" spatial attention respectively. Overt attention occurs when an animal chooses to make a small eye movement, or saccade, to a region. Covert attention refers to the ability to increase visual information processing of a region without an eye movement. The challenge for machine learning when trying to use these forms of spatial attention is to effectively choose which regions of the image on which to focus attention.

Feature-based attention is a kind of covert attention in that it does not require eye movements. However, it serves to enhance certain features of an image, rather than certain locations. Some work in CNNs has approximated this style of attention. Most prominently, the work of \cite{stollenga2014deep} finds a policy for dynamically weighting feature maps in a Maxout network. While powerful, this technique requires iterative training and many feedforward-feedback loops at runtime. The implementation of FBA here is applied to a specific object category in order to enhance detection of that category, and only requires one feedforward pass.    

\subsection{Findings from Biology}
Attention has been found to increase performance accuracy and decrease reaction time on a variety of cognitive tasks. The neural mechanisms underlying covert attention are a hot topic of research in neuroscience. Much of this work comes from non-human primate studies, wherein neural activity is recorded from various areas in the ventral visual stream as visual stimuli are presented to the animal. These studies have lead to the feature similarity gain model of attention \citep{treue1999feature}. It states that when an animal is cued to attend to a certain visual feature, neurons that are selective to that feature increase their firing rate beyond the rate found without attention \citep{maunsell2006feature}. In most studies, this increase is found to be a multiplicative effect. Furthermore, effects are stronger in later areas of the ventral stream \citep{mcadams1999effects}. 

Some studies have shown that neurons that are selective to features other than those attended have their activity suppressed. This represents a bi-directional modulation of neural activity with attention. Some studies suggest changes in the positive direction are more prevalent, however \citep{mcadams1999effects}.

Two lines of results show that this attention is spatially global and feature specific: (1) FBA alters the activity of neurons selective to the attended feature at all spatial locations across both hemispheres \citep{cohen2011using} and (2) In tasks with attention applied to one of two spatially overlapping stimuli, neural activity is biased toward the activity found when the attended feature is presented alone \citep{patzwahl2009combining}.

\begin{figure}

\includegraphics[width=1\linewidth]{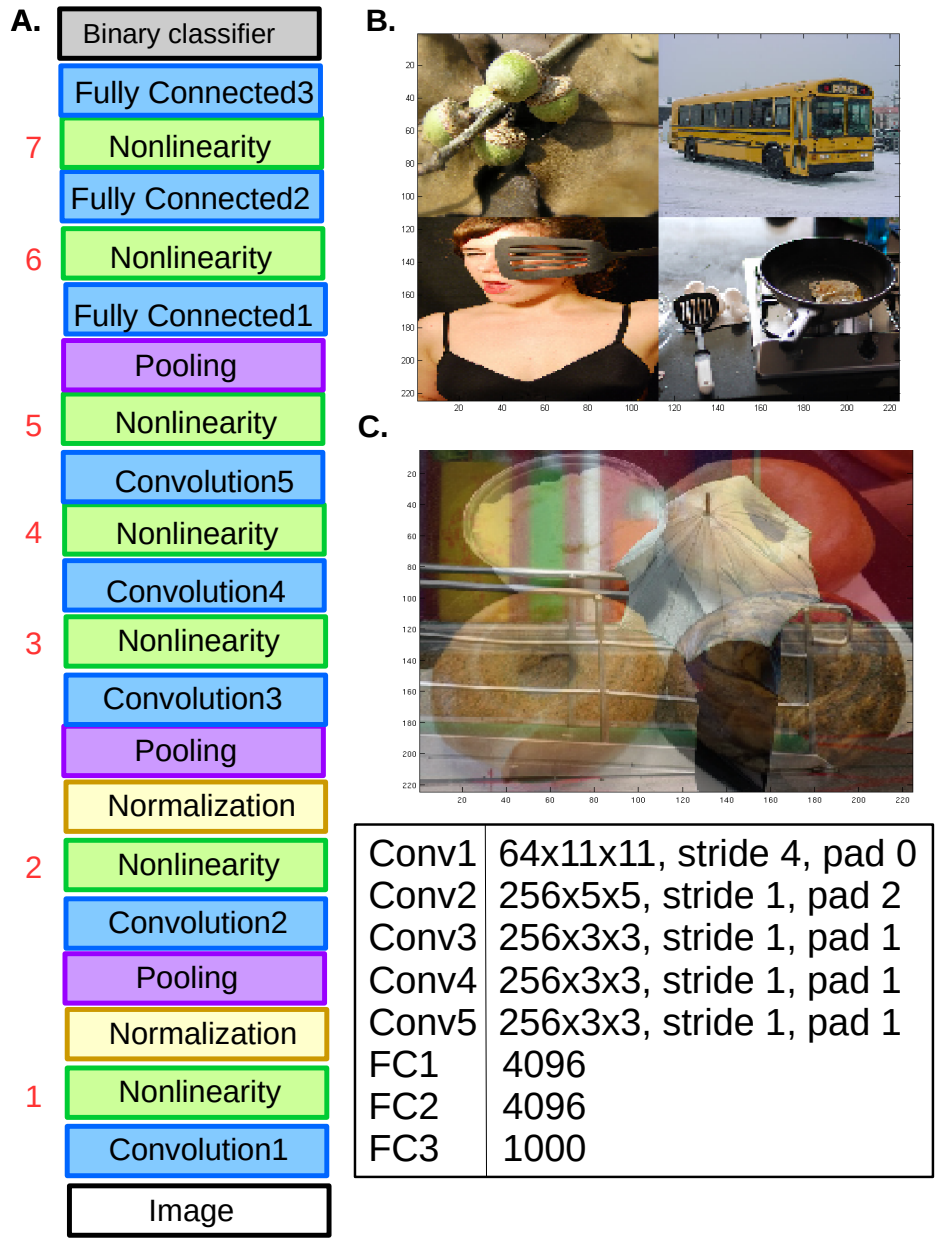}
\caption{Architecture and task. (A) The architecture of the pre-trained CNN. For the object detection task, the 1000-way softmax classifier normally used in the CNN is replaced by binary classifiers trained to determine if a given object category is present or absent. Category-specific feature-based attention can be applied at any of the ReLU layers, marked with the red numbers on the side. Filter details are given in the box. Special imagesets were made to test the object detection abilities of FBA. Array images (B) are composed of 4 ImageNet images on a 2x2 grid. Merged images (C) are two overlaid ImageNet images. }
\label{arch}
\end{figure}

\section{FBA Implementation and Testing}
Taking inspiration from biology, I've incorporated into a CNN the mechanisms used by neurons for attending to specific features or objects. This allows enhanced object detection. Different implementation details are described below and tested with a pre-trained CNN on two types of object detection tasks . 

\subsection{FBA Procedure}
FBA works by enhancing the features of an attended object category. This can be implemented in a pre-trained CNN by altering the network in a category-specific way. In order to determine how different feature maps should be altered for a given category, information about their average activity in response to that category is collected. When processing a new image with FBA, feature maps are altered to be biased in the direction of the average activity of the attended category.    

\subsubsection{Determination of Feature Patterns}
Category-specific \textit{feature patterns} are created based on the average activity of feature maps when presented with images of a given category. Feature patterns are defined for each category and for each ReLU layer in the network (which in the network used here, shown in Figure \ref{arch}, corresponds to layers 2,6,10,13,16,19,21. These layers are marked with red numbers that will be used to refer to them in further figures).

Activity of the $k^{th}$ feature map in layer $l$, in response to image $n$, is given as $\mathbf{X}_{lk}(n)$, with $\mathbf{X}_{lk} \in \mathbb{R}^{h \times w}$. This activity can be averaged over the two spatial dimensions to give a scalar value, $r_{lk}(n)$:
\begin{equation}
r_{lk}(n)=\frac{1}{hw}\sum_{i=1}^h \sum_{j=1}^w x^{ij}_{lk}(n)
\end{equation}
where $h$ and $w$ are the height and width of the feature map, respectively, and $x^{ij}_{lk}$ is the $ij^{th}$ element of $\mathbf{X}_{lk}$. (For the fully connected layers, which lack two-dimensional feature maps, $r_{lk}(n)=x_{lk}(n)$, which is just the activity of an individual node). Thus, the $k^{th}$ element of the vector $\mathbf{r}_{l}(n)$, is the spatially-averaged activity of the $k^{th}$ feature map in response to image $n$. Averaging these values over all images in the training set gives the vector $\bar{\mathbf{r}_l}$:    
\begin{equation}
\bar{\mathbf{r}_l}=\frac{1}{N}\sum_{n=1}^N \mathbf{r}_l(n)
\end{equation}
Next, feature patterns are defined for each layer, $l$, and object category, $c$. A feature pattern, given as $\mathbf{f}_l^c$, is a vector with an entry for each feature map, which determines how that feature map is altered when attention is applied to category $c$. It is defined as:
\begin{equation}
\mathbf{f}_l^c=\frac{\frac{1}{N_c} \sum_{n\in c}\mathbf{r}_l(n)-\bar{\mathbf{r}_l}}{\sqrt{\frac{1}{N} \sum_{i=1}^N(\mathbf{r}_l(n)-\bar{\mathbf{r}_l})^2}}
\end{equation}

with $N_c$ representing the total number of training images from object category $c$. That is, an object category's feature pattern at a given layer is merely the average activity of the feature maps at that layer in response to images of that category, with the mean activity under all image categories subtracted and standard deviation divided. These feature patterns determine how the feature maps are modulated when attention is applied to a specific category.    

At this point, a choice can be made about whether to set negative elements of these feature patterns to zero. As discussed above, evidence suggests that neurons may be modulated both positively and negatively by attention, but positive modulation may be stronger. Both rectification options are tested here.  

\subsubsection{Options for Application of FBA at Runtime}
Once the feature patterns are generated, they are used to alter the network at runtime. Here again there are many implementation options.

First, the alteration can manifest as either an additive effect or a multiplicative one. That is, when attending to category $c$, a weighted version of the feature pattern for category $c$ can be added before the rectified linear units:
\begin{equation}
\label{add}
x_{lk}^{ij}=ReLU(I_{lk}^{ij}+\beta f^c_{lk})
\end{equation}
with $I_{lk}^{ij}$ representing input to the ReLU coming from layer $l-1$.   
Or, for the multiplicative effect, the slope of the rectified linear units can be multiplied by a weighted function of the feature pattern for category $c$:
\begin{equation}
\label{mult}
x_{lk}^{ij}=(1+\beta f^c_{lk})ReLU(I_{lk}^{ij})
\end{equation}
Strength of the attention is varied via the weighting parameter, $\beta$. For the additive effect, $\beta$ was varied from 4 to 24, and for the multiplicative effect $\beta$ was varied from .2 to 1.2. These values were found to give a range of performances. 

Combining the modulation options described in \ref{add} and \ref{mult} with the rectification options discussed above, four implementation combinations are possible: Additive-Bidirectional, Additive-Positive, Multiplicative-Bidirectional, and Multiplicative-Positive.

Finally, attention does not need to be applied to all layers. As discussed above, the strongest effects of attention are found in later extrastriate visual areas. Thus, performance is measured here when attention is applied to different layers individually, and to combinations of layers.

In all cases, the modulation is applied in a spatially global way. That is, each $ij$ position in a feature map receives the same modulation.  

\subsection{Object Detection Tests}
To show how FBA can enhance object detection, two image sets were created which pose difficult problems for a standard CNN. These image sets are described below. In either case, the task of the network was to output a binary variable reporting whether or not a given object category was present in the image. For this, the softmax classifier is removed and the last layer of the network was fed into a binary classifier trained to detect the given category (results from SVM shown, though results are similar using regularized logistic regression). The test images are described here:

\subsubsection{Array Imageset}
Array images are each composed of four ImageNet images selected randomly from the available categories, arranged in a 2x2 grid (example in Figure \ref{arch}). This type of "cluttered" image is reminiscent of those used in experiments on feature and spatial attention.  
\subsubsection{Merged Imageset}
Merged images are each composed of two ImageNet images superimposed on top of one another as a weighted linear sum of pixel values (example in Figure \ref{arch}). This type of image is an analogue of the stimuli used in \cite{patzwahl2009combining}. It offers a test of FBA's ability to "see" an object category when distracting features are overlapping. More generically, the overlaid image could simply be viewed as a form of structured noise.   

\subsection{Network Architecture and Parameters}
The pre-trained network used for this work comes from \cite{vedaldi15matconvnet}. Its architecture and details are shown in Figure \ref{arch}. It contains 7 ReLU layers, and thus 7 possible locations of attention.

For each category, a binary classifier was trained on 150 category and 150 non-category normal images from ImageNet (results do not depend strongly on number of training images). Tests of performance shown here were on the same number of test images coming from the above test imagesets. Training images do not overlap with the images used to make the imagesets. Performance is determined by averaging over 20 folds of training on different subsets of training images. This distribution of performances allows for significance testing between different implementation options.

\section{Results}
\begin{figure}[ht]
\includegraphics[width=1\linewidth]{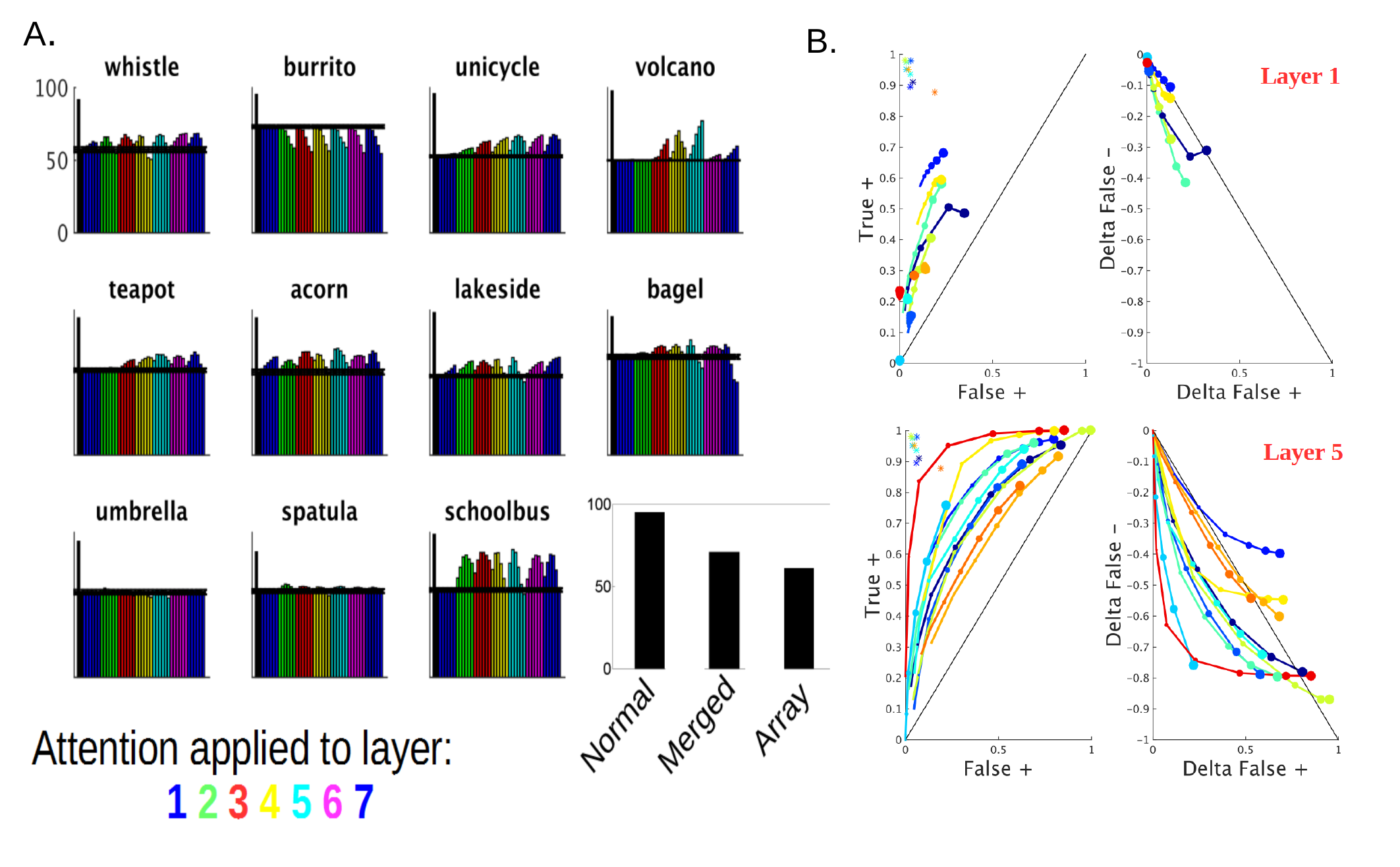}
\caption{Effects of FBA on different object categories, applied at different layers with various strengths. A.) Black vertical line represents performance on normal images for each category. Black horizontal line represents performance on array images, with width of one standard deviation. Different colors show the effect of FBA applied at different individual layers (1-7) and the bars within a color reflect increasing strength. The implementation options used here are multiplicative \& bi-directional effects. The inset shows the average binary classification performance for merged and array images compared to normal images (classifier trained on normal images). B.) The effect of attention strength (applied to Layers 1 and 5) on true \& false positives and negatives. Different colors correspond to different categories and increasing dot size is increasing strength. Left column is a standard ROC curve and asterisks show performance on normal images. Right column shows how the false positive and false negative rates change from baseline (i.e., no attention) with increasing attention strength    }
\label{bars1}
\end{figure}

The array and merged imagesets prove challenging for the binary classifiers trained on normal images. As the inset in Figure \ref{bars1} shows, the binary classification performance (averaged over all categories) is high on normal images (95.17\%), but is lower for merged images (70.89\%) and even lower for array images(59.29\%). More standard methods of assessing CNNs also show this difficulty: top-5 error rate on merged images (calculated by accepting either of the two image categories in the merged image as correct) is 58.18\%.

Applying FBA, however, increases the ability of the network to detect the attended object. Different implementation options provide different results. Results are also dependent on the category attended. Figures shown are of results from array images, but merged image results are qualitatively similar and shown in the Appendix. Array images were more challenging, and show larger increases in performance.   

\begin{figure}[ht]

\includegraphics[width=.75\linewidth]{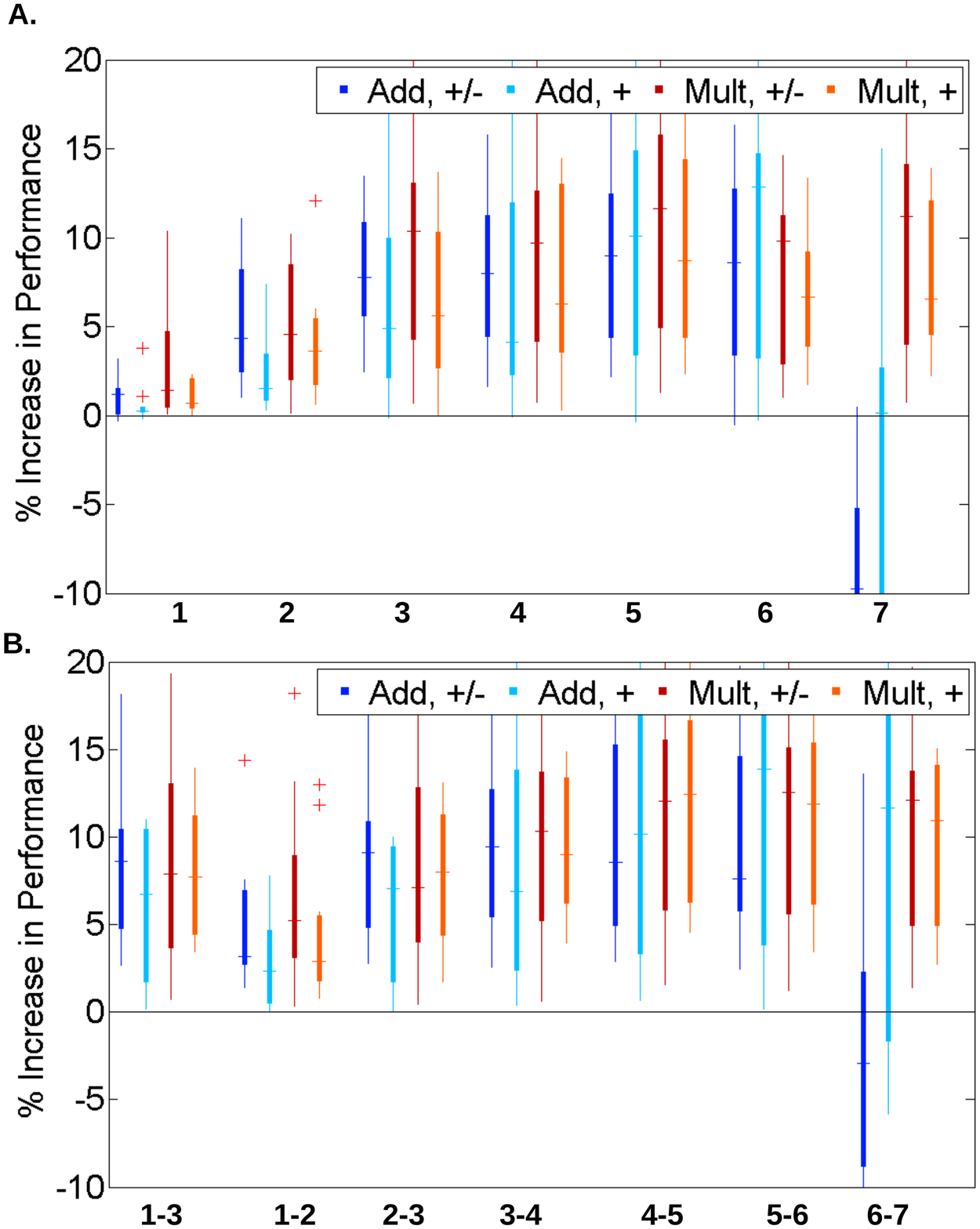}
\caption{Increase in binary classification performance for different FBA implementation options at different single layers  (A) and combinations of layers (B), as labeled on the X axes. The Y axis represents performance with attention minus performance without, resulting in a measure of the increase in performance in units of percentage points. Horizontal lines represent median performance increase across categories. Boxes are 25th and 75th percentiles and whiskers extend to the most extreme data points not considered outliers. Red crosses are outliers. The best mean performance increase for the single layer application is 12.49\% (layer 5, multiplicative-bidirectional). The best mean performance increase for multiple layer application is 13.16\% (layers 4-5, multiplicative-positive). Data shown here for array images, merged image data in Appenix Figure \ref{bwpM} }
\label{bwp}
\end{figure}

\subsection{Strength of Attention Affects Performance}
Figure \ref{bars1} shows the effect of different strengths ($\beta$) of attention when it is applied at different layers (the implementation options used here are multiplicative bi-directional effects, tested on array images). In many instances, increasing the strength of attention is beneficial to a point, and then becomes detrimental to performance. This can be shown by tracing the effect of increasing attention strength through a space of changing true and false negative rates, as seen in the right column of Figure \ref{bars1}B. In order for FBA to be effective, it should decrease false negatives without substantially increasing false positives. As can be seen, for most categories (shown in different colors), false-negative decrease rate outpaces false positive increase rate, up to a certain strength (increasing strength shown as increasing circle size). The left column of Figure \ref{bars1}B show ROC curves that result from varying strength. Again, increases in strength bring performance closer to that found on the normal images (asterisks), but only to a point. Notably, the effects of increasing strength are different for different layers. Earlier layers don't move the performance as much, even at high strengths, compared to later layers.   

For further analyses, the best performing strength at a given layer and category will be used.

\subsection{Applying FBA to Later Layers Performs Best}
As Figure \ref{bwp}A shows, the average increase in performance (in units of percentage points) with attention across categories varies based on the layer at which attention is applied. Looking at median performance, applying attention to later layers (specifically 5 and 6) has the best effect on performance across implementation details. Interestingly, applying additive FBA to the 7th layer causes an average decrease in performance. 

Applying FBA to multiple layers at once (Figure (\ref{bwp}B, half-strength was used when applied to multiple layers at once) appears to have only minor effects on performance compared to single layers. 

Averaging over categories can make seeing the impact of applying attention to different layers difficult. In order to see this impact, a direct comparison was made. For each category, performance was directly compared across layers and the layer that led to the best performance was found. This was done for all four combinations of implementation options. As shown in Figure \ref{hists}C, layer 5 consistently outperforms other layers. 

\subsection{Bi-directional Multiplicative Effects Perform Best}
In order to concretely determine which of the different implementation options performs best, a direct comparison was made, similar to the one described above for Figure \ref{hists}C. That is, at each combination of category and layer, the four pairings of implementation options were compared and the best performing pairing was determined. Figure \ref{hists}B shows histograms displaying the results of these comparisons. The combination of bi-directional and multiplicative effects clearly outperforms the other options.  

\subsection{Effects of Bi-directional Modulation Are Unclear}
\begin{figure}

\includegraphics[width=.9\linewidth]{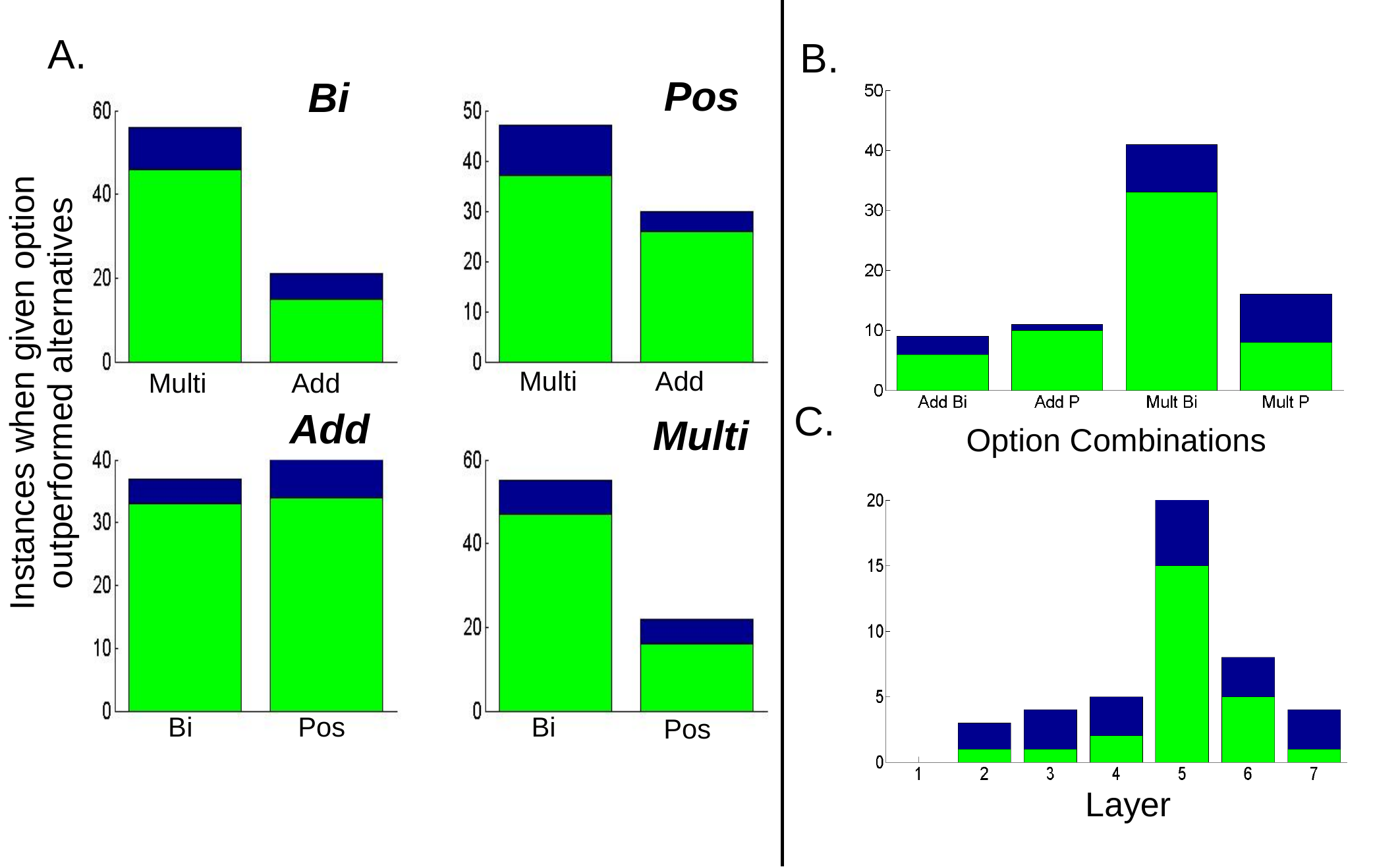}
\caption{Histograms of instances when one implementation option performs better than others. Blue bars show number of instances when the given option leads to a larger increase in performance than other option(s). Green bars show number of times when that difference is statistically significant. (A) Direct comparisons of different implementation options. In the upper left (right), the comparison is made between multiplicative effects and additive effects under conditions when the modulation is bi-directional (positive). In the bottom left (right) bi-directional and positive modulations are compared when effects are additive (multiplicative). Data shown here is for array images. Merged image data is in Appendix Figure \ref{histosM} }
\label{hists}
\end{figure}

Breaking down these comparisons even further shows an interesting result (Figure \ref{hists}A).  

\begin{figure}[t]

\includegraphics[width=1\linewidth]{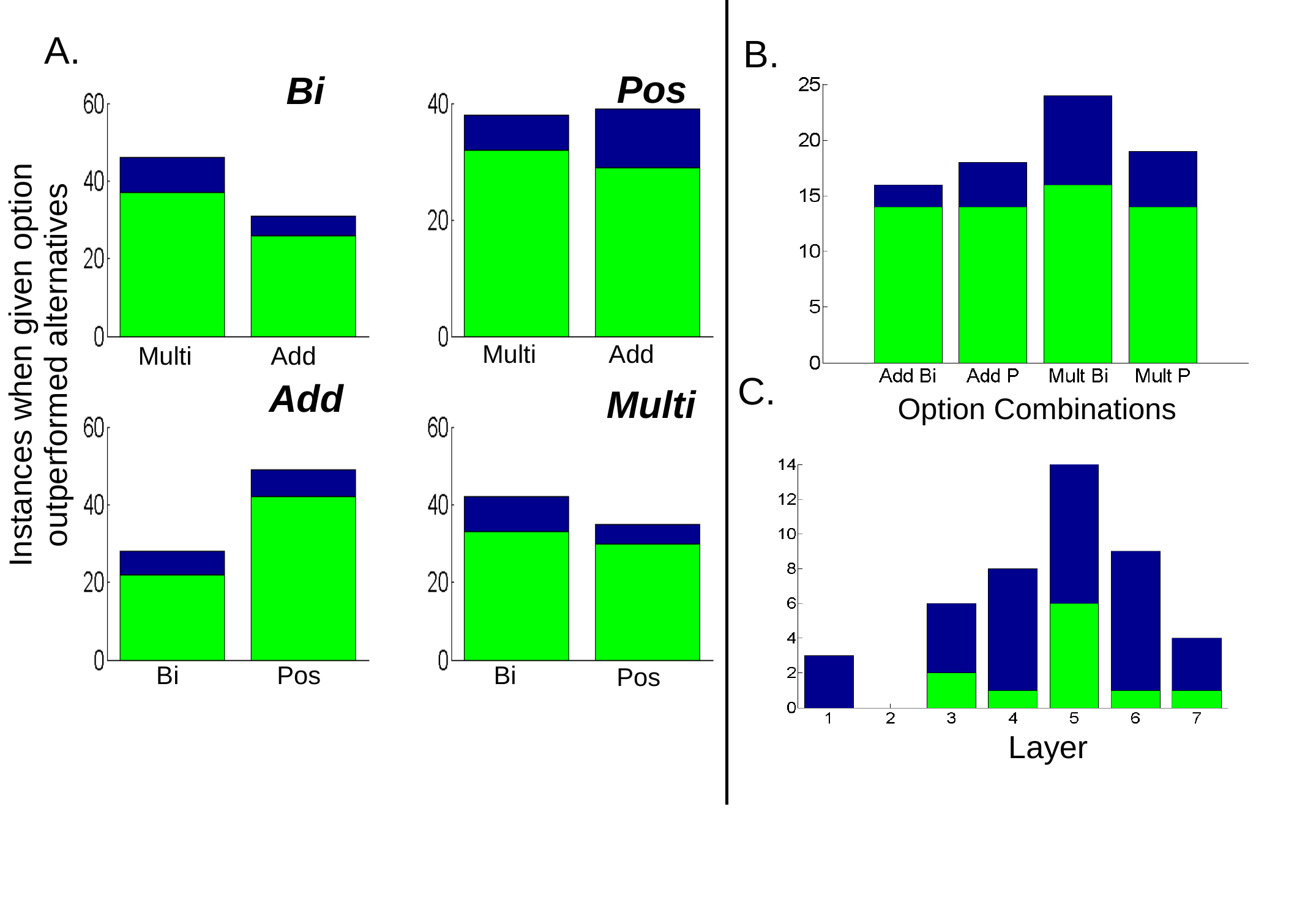}
\caption{Same as Figure \ref{hists}, but for merged image data.}
\label{histosM}
\end{figure}

Looking at all instances when multiplicative effects were used, it's clear that bi-directional modulation performs better than positive-only. However when looking at instances where additive effects are used, there is not a clear winner (or, in the case of merged image data (Appendix Figure \ref{histosM}), positive modulation is the winner). Thus, there appears to be a synergistic benefit from combining multiplicative effects and bi-directional modulation. 

Under this investigation, the benefit of multiplicative effects remains clear. That is, looking at all instances when bi-directional modulation is used, it's clear that multiplicative effects perform better than additive, and the same is true when looking at positive-only modulation (Figure \ref{hists}A, top row).

\subsection{Complications}
While FBA does increase performance on binary object detection tasks, there are some complications. First, not all categories are affected equally. This is perhaps due to the style of training images used to make the feature patterns for each category. Images in categories that perform well, like schoolbus, tend to have the object centered, and from somewhat consistent angles. Thus, learned feature patterns for these categories strongly represent the object. Food images, such as bagel and burrito, tended to be cluttered and display the object in multiple different ways, weakening the ability of the feature patterns to capture the relevant traits.

Another difficulty comes from determining the best strength to use. Here, strength was treated as a free parameter and the best value in each instance was experimentally determined. Generally, the best strength to use will depend on the difficulty of the imagesets being evaluated. Biologically, the brain must have a way of setting the strength of the feedback connections that control attention.

\section{Conclusion}
\begin{figure}[t]

\includegraphics[width=1\linewidth]{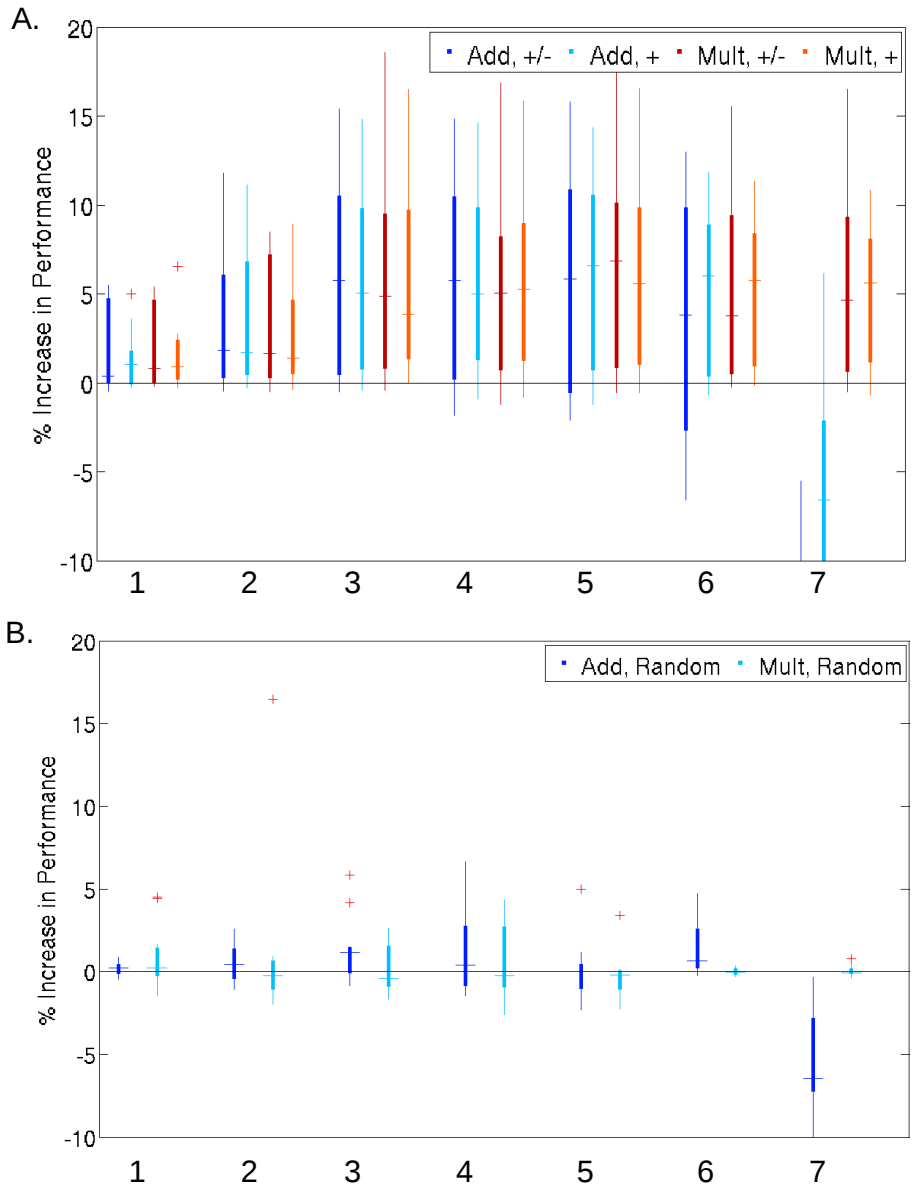}
\caption{(A) Same as Figure \ref{histosM} (data is from merged images), however, feature patterns are determined by averaging over the activity from only 20 images per category, as opposed to the 150 used in other figures. This shows that FBA can increase performance even when using relatively little data. Furthermore, these increases are not trivial, as random perturbations of feature patterns are not capable of achieving them: (B) shows performance for one instantiation of randomly perturbed feature patterns. }
\label{QRbwp}
\end{figure}

The implementation of FBA presented here is a simple feedforward operation that does not require iterative training. It allows for the ability to train on normal, clear images and test on more challenging images. As such, it may serve as a means of generalizing the classification ability of a CNN. A nice feature of the FBA implementations described here is that even when little data (20 images for each category) is used to make the feature patterns, performance still increases substantially (Appendix Figure \ref{QRbwp}).   

Although not tested here, FBA may also aid in fine discrimination tasks, which is another setting where attention is used by humans. The freedom to apply FBA at different layers may be especially useful for fine discrimination problems, as lower layers represent the finer, smaller features.  

Aside from performance, this work further demonstrates the applicability of biological ideas to CNNs. And conversely, the ability to test out biological mechansims in CNNs. This work provides evidence that FBA applied as a multiplicative effect to later layers in the visual stream is an effective way to increase performance, more so than additive or lower layer effects. Thus, biology appears to be using the most effective option for increasing visual information processing under attention. 

\subsubsection*{Acknowledgments}

Thanks to Ken Miller and Josh Merel for input on this project. This work was done at the Center for Theoretical Neuroscience, with funding from the Kavli Institute, Gatsby Charitable Foundation, Schwartz Foundation, and the Zuckerman Mind Brain Behavior Institute. 

\bibliography{iclr2016_conference}

\begin{thebibliography}{10}
\providecommand{\natexlab}[1]{#1}
\providecommand{\url}[1]{\texttt{#1}}
\expandafter\ifx\csname urlstyle\endcsname\relax
  \providecommand{\doi}[1]{doi: #1}\else
  \providecommand{\doi}{doi: \begingroup \urlstyle{rm}\Url}\fi

\bibitem[Cohen \& Maunsell(2011)Cohen and Maunsell]{cohen2011using}
Cohen, Marlene~R and Maunsell, John~HR.
\newblock Using neuronal populations to study the mechanisms underlying spatial
  and feature attention.
\newblock \emph{Neuron}, 70\penalty0 (6):\penalty0 1192--1204, 2011.

\bibitem[Maunsell \& Treue(2006)Maunsell and Treue]{maunsell2006feature}
Maunsell, John~HR and Treue, Stefan.
\newblock Feature-based attention in visual cortex.
\newblock \emph{Trends in neurosciences}, 29\penalty0 (6):\penalty0 317--322,
  2006.

\bibitem[McAdams \& Maunsell(1999)McAdams and Maunsell]{mcadams1999effects}
McAdams, Carrie~J and Maunsell, John~HR.
\newblock Effects of attention on orientation-tuning functions of single
  neurons in macaque cortical area v4.
\newblock \emph{The Journal of Neuroscience}, 19\penalty0 (1):\penalty0
  431--441, 1999.

\bibitem[Mnih et~al.(2014)Mnih, Heess, Graves, et~al.]{mnih2014recurrent}
Mnih, Volodymyr, Heess, Nicolas, Graves, Alex, et~al.
\newblock Recurrent models of visual attention.
\newblock In \emph{Advances in Neural Information Processing Systems}, pp.\
  2204--2212, 2014.

\bibitem[Patzwahl \& Treue(2009)Patzwahl and Treue]{patzwahl2009combining}
Patzwahl, Dieter~R and Treue, Stefan.
\newblock Combining spatial and feature-based attention within the receptive
  field of mt neurons.
\newblock \emph{Vision research}, 49\penalty0 (10):\penalty0 1188--1193, 2009.

\bibitem[Stollenga et~al.(2014)Stollenga, Masci, Gomez, and
  Schmidhuber]{stollenga2014deep}
Stollenga, Marijn~F, Masci, Jonathan, Gomez, Faustino, and Schmidhuber,
  J{\"u}rgen.
\newblock Deep networks with internal selective attention through feedback
  connections.
\newblock In \emph{Advances in Neural Information Processing Systems}, pp.\
  3545--3553, 2014.

\bibitem[Treue \& Trujillo(1999)Treue and Trujillo]{treue1999feature}
Treue, Stefan and Trujillo, Julio C~Martinez.
\newblock Feature-based attention influences motion processing gain in macaque
  visual cortex.
\newblock \emph{Nature}, 399\penalty0 (6736):\penalty0 575--579, 1999.

\bibitem[Vedaldi \& Lenc()Vedaldi and Lenc]{vedaldi15matconvnet}
Vedaldi, A. and Lenc, K.
\newblock Matconvnet -- convolutional neural networks for matlab.

\bibitem[Xu et~al.(2015)Xu, Ba, Kiros, Courville, Salakhutdinov, Zemel, and
  Bengio]{xu2015show}
Xu, Kelvin, Ba, Jimmy, Kiros, Ryan, Courville, Aaron, Salakhutdinov, Ruslan,
  Zemel, Richard, and Bengio, Yoshua.
\newblock Show, attend and tell: Neural image caption generation with visual
  attention.
\newblock \emph{arXiv preprint arXiv:1502.03044}, 2015.

\bibitem[Zeiler \& Fergus(2014)Zeiler and Fergus]{zeiler2014visualizing}
Zeiler, Matthew~D and Fergus, Rob.
\newblock Visualizing and understanding convolutional networks.
\newblock In \emph{Computer Vision--ECCV 2014}, pp.\  818--833. Springer, 2014.

\end{thebibliography}
\bibliographystyle{iclr2016_conference}

\pagebreak

\section{Appendix}
This section contain supplementary figures.
\begin{figure}[h]

\includegraphics[width=.9\linewidth]{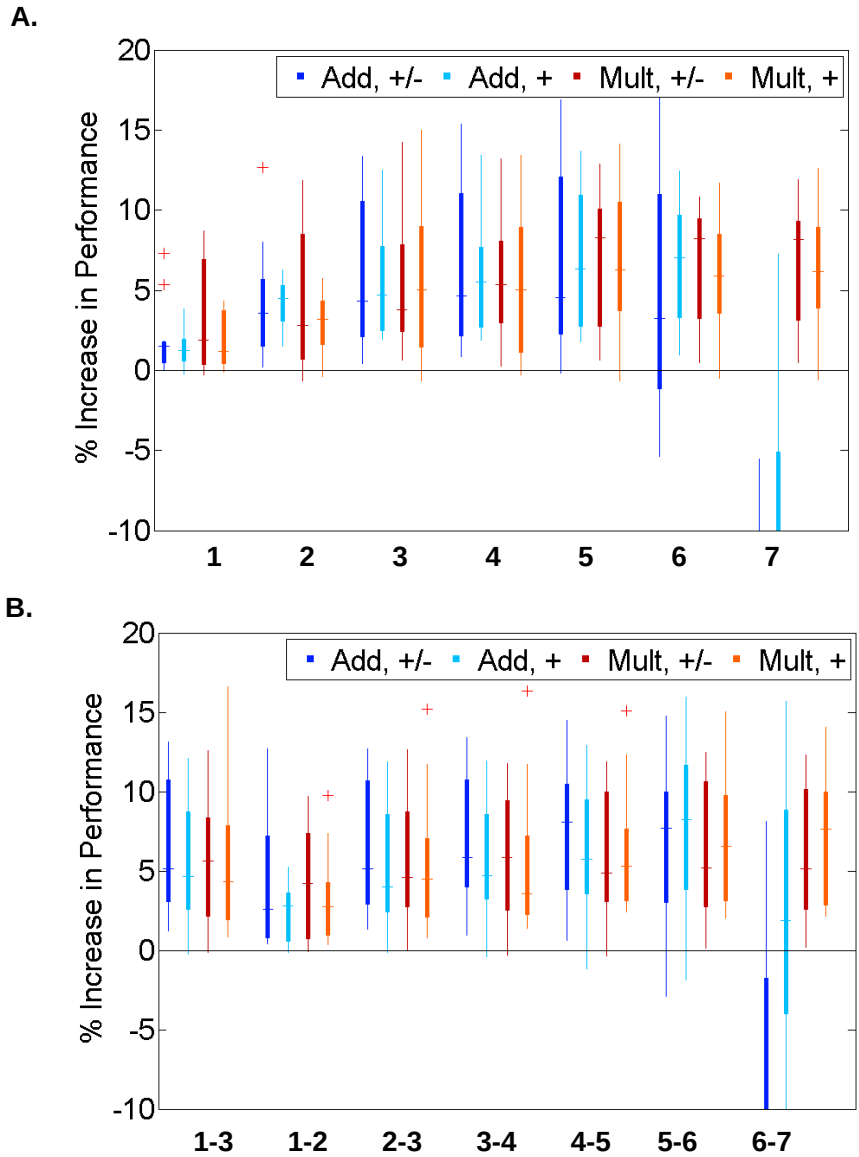}
\caption{Same as Figure \ref{bwp}, but for merged image data.}
\label{bwpM}
\end{figure}

\end{document}